\documentclass[10pt,twocolumn,letterpaper]{article}

\usepackage[pagenumbers]{cvpr} % To force page numbers, e.g. for an arXiv version

\usepackage[accsupp]{axessibility} % Improves PDF readability for those with visual impairments.
\usepackage{amssymb}
\usepackage{bbding}
\usepackage{amsmath}
\usepackage{makecell}
\usepackage{makebox}
\usepackage{graphicx}
\usepackage{array}
\usepackage{svg}
\usepackage{multirow}
\usepackage{float}
\usepackage{appendix}

\definecolor{cvprblue}{rgb}{0.21,0.49,0.74}
\usepackage[pagebackref,breaklinks,colorlinks,citecolor=cvprblue]{hyperref}

\def\paperID{5344} % *** Enter the Paper ID here
\def\confName{CVPR}
\def\confYear{2024}

\title{Unleashing the Potential of SAM for Medical Adaptation \\ via Hierarchical Decoding}

\author{Zhiheng Cheng\textsuperscript{1} \quad Qingyue Wei\textsuperscript{2} \quad Hongru Zhu\textsuperscript{3} \quad Yan Wang\textsuperscript{1} \quad Liangqiong Qu\textsuperscript{4} \quad 
\\Wei Shao\textsuperscript{5} \quad Yuyin Zhou\textsuperscript{6} \vspace{.3em}\\ 
\normalsize{\textsuperscript{1}Shanghai Key Laboratory of Multidimensional Information Processing, East China Normal University}\\
\normalsize{\textsuperscript{2}Stanford University} \quad
\normalsize{\textsuperscript{3}Johns Hopkins University}\quad
\normalsize{\textsuperscript{4}University of Hong Kong}\\
\normalsize{\textsuperscript{5}University of Florida}\quad
\normalsize{\textsuperscript{6}UC Santa Cruz}
\vspace{-1em}
}
\begin{document}
\maketitle
\begin{abstract}
The Segment Anything Model (SAM) has garnered significant attention for its versatile segmentation abilities and intuitive prompt-based interface. However, its application in medical imaging presents challenges, requiring either substantial training costs and extensive medical datasets for full model fine-tuning or high-quality prompts for optimal performance.
This paper introduces H-SAM: a prompt-free adaptation of SAM tailored for efficient fine-tuning of medical images via a two-stage hierarchical decoding procedure. In the initial stage, H-SAM employs SAM's original decoder to generate a prior probabilistic mask, guiding a more intricate decoding process in the second stage. Specifically, we propose two key designs: 1) A class-balanced, mask-guided self-attention mechanism addressing the unbalanced label distribution, enhancing image embedding; 2) A learnable mask cross-attention mechanism spatially modulating the interplay among different image regions based on the prior mask. 
Moreover, the inclusion of a hierarchical pixel decoder in H-SAM enhances its proficiency in capturing fine-grained and localized details. 
This approach enables SAM to effectively integrate learned medical priors, facilitating enhanced adaptation for medical image segmentation with limited samples.
Our H-SAM demonstrates a 4.78\% improvement in average Dice compared to existing prompt-free SAM variants for multi-organ segmentation using only 10\% of 2D slices.
Notably, without using any unlabeled data, H-SAM even outperforms state-of-the-art semi-supervised models relying on extensive unlabeled training data across various medical datasets. Our code is available at ~\url{https://github.com/Cccccczh404/H-SAM}.
\end{abstract}    
\begin{figure}
  \centering
   \includegraphics[width=\linewidth]{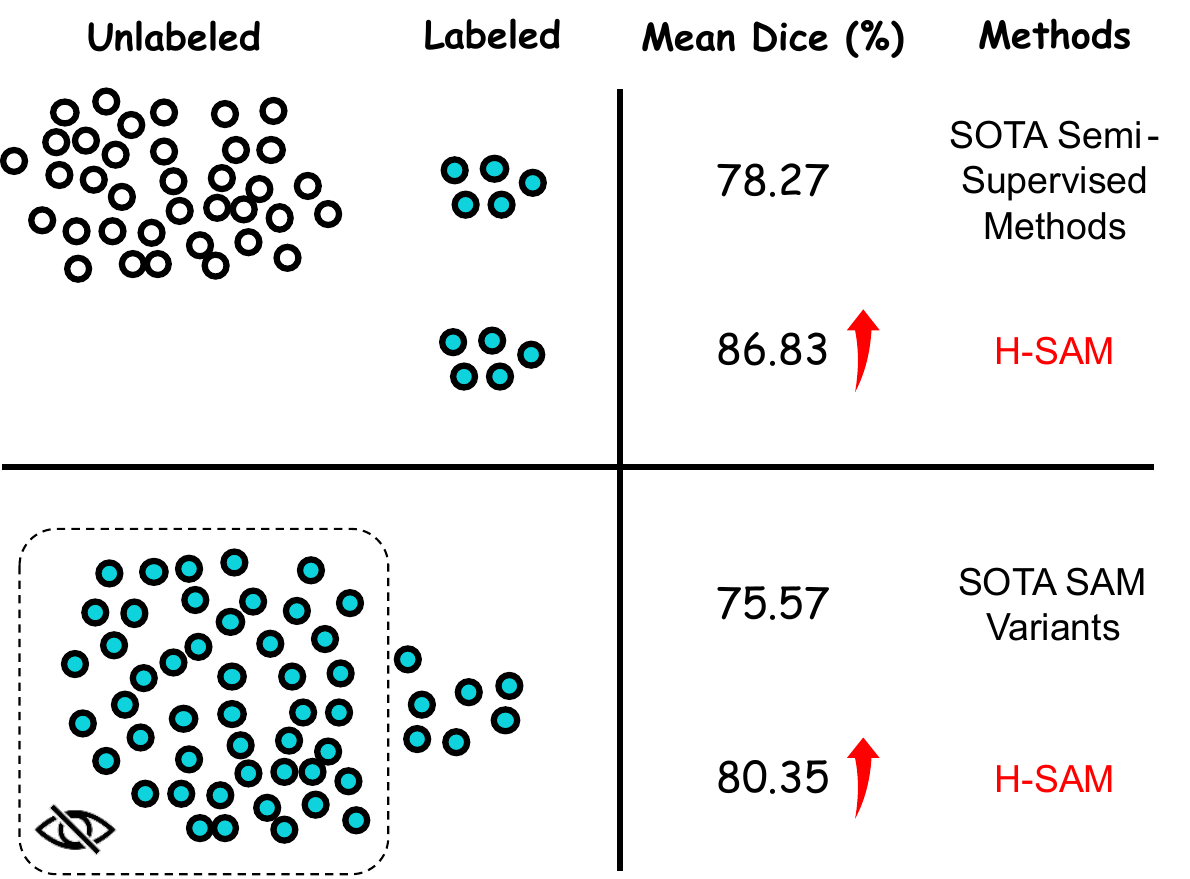}

   \caption{H-SAM is advantageous in few-shot medical image segmentation. It achieves over 80\% in average Dice using only 10\% slices for multi-organ segmentation, outperforming existing prompt-free SAM adaptation methods. Without using any unlabeled data at all, it even outperforms state-of-the-art semi-supervised models that use extensive unlabeled training data for prostate segmentation. }
   \label{fig1}
\end{figure}

\section{Introduction}
\label{sec:intro}

Accurate delineation of tissues, organs, and regions of interest through medical image segmentation is pivotal in aiding medical professionals' diagnostic precision and treatment planning processes~\cite{cheng2022resganet,de2018clinically}. Furthermore, it plays a fundamental role in propelling disease research and discovery~\cite{ouyang2020video}. Nonetheless, a significant challenge in this field lies in the demand for deep learning models to undergo extensive training on large annotated datasets, a resource often challenging to procure within the medical domain.

Recently, Segment Anything Model (SAM) \cite{kirillov2023segment}, trained with over a billion masks from diverse natural images, demonstrates remarkable zero-shot learning capabilities. This breakthrough presents an avenue for significant advancements in medical image segmentation, especially considering the limited availability of extensive datasets in the medical realm. However, SAM's performance on medical images diminishes notably in zero-shot settings, exhibiting reduced accuracy and robustness~\cite{mazurowski2023segment, he2023accuracy, li2023polyp, hu2023skinsam, ji2023sam}. This decline can be attributed to SAM's lack of exposure to medical images during training, as its extensive training revolves around natural images \cite{huang2023segment,zhang2023segment}. 

While training SAM exclusively on medical datasets is a potential solution, it incurs substantial training costs and risks of over-fitting to single datasets \cite{ma2023segment}. Efforts to bridge the gap between medical and natural image domains involve adapting SAM to specific medical datasets \cite{gao2023desam, cui2023all, zhang2023sam}. Previous works primarily focus on inserting adapter layers into the image encoder with minimal decoder changes~\cite{zhang2023customized,gao2023desam}. Most of these efforts employ prompted SAM adaptation, generating prompts using point or bounding boxes from ground truth during testing~\cite{feng2023cheap,wu2023medical,zhang2023self}. 
However, creating accurate prompts demands domain knowledge from medical experts, which is often limited, time-consuming, and prone to noise, compromising segmentation accuracy.
In response, prompt-free SAM adaptation methods~\cite{zhang2023customized,chen2023sam,rahman2023g} have emerged, yet they typically yield inferior results compared to prompted methods due to the lack of medical knowledge that prompts provide.

\begin{figure*}
  \centering
   \includegraphics[width=\linewidth]{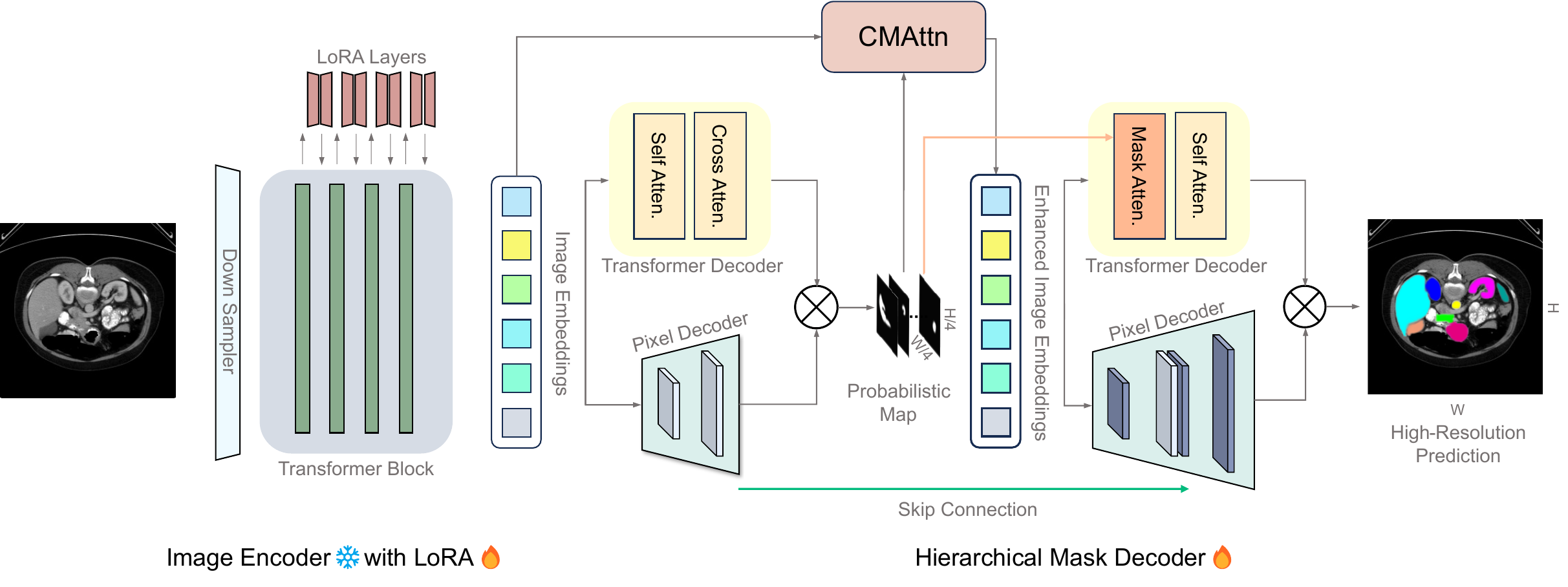}
   \vspace{-1.5em}
   \caption{The H-SAM framework integrates a LoRA-adapted image encoder and a sophisticated 2-stage hierarchical decoder. We finetune the prompt encoder with default embeddings under a prompt-free setting. A key innovation lies in our hierarchical mask decoder, which strategically utilizes predictions from the stage-1 decoder as priors to achieve nuanced segmentation with 2 implementations: Class-Balanced Mask-Guided Self-Attention (CMAttn), and Learnable Mask Cross-Attention. And a hierarchical pixel decoder is employed to complement the enriched object queries derived from the transformer decoder.}
   \label{framework}
\end{figure*}

We present H-SAM, a prompt-free variant of the Segment Anything Model (SAM), aimed at integrating medical knowledge via a streamlined two-stage hierarchical mask decoder while maintaining the image encoder frozen.
Initially, input images are processed by a LoRA-adapted image encoder. H-SAM employs SAM's original lightweight mask decoder in the first stage to generate a prior probabilistic mask, guiding a more intricate second decoding stage. 
Two key designs underpin this process: 
1)  A \textbf{class-balanced, mask-guided self-attention} mechanism recalibrates the image embedding using self-attention from the prior mask, ensuring balanced representation across categories with noise augmentation.
2) A \textbf{learnable mask cross-attention} mechanism employs the prior mask to modulate cross-attention spatially within the subsequent Transformer decoder, attenuating less relevant background noise.
Moreover, a hierarchical pixel decoder complements the hierarchical Transformer decoder, enhancing the model's precision and ability to capture localized details. Figure~\ref{framework} illustrates the overall pipeline.

In both fully-supervised and few-shot multi-organ segmentation tasks, our H-SAM surpasses existing prompt-free SAM variants. Specifically, using 10\% and 100\% of 2D slices, H-SAM achieves an average Dice score improvement of 4.78\% and 3.48\% on the Synapse dataset, respectively. 
This superior performance is also evident in other few-shot medical image segmentation tasks including the left atrial (LA) dataset and PROMISE2012 dataset.
Notably, as illustrated in Figure~\ref{fig1}, H-SAM excels without the need for any unlabeled data, surpassing state-of-the-art semi-supervised models that rely on extensive unlabeled training datasets. The promising performance of $87.27\%$ and $89.22\%$ achieved on the prostate and left atrial segmentation using only 3 and 4 cases highlights H-SAM's potential in medical imaging applications.

\section{Related Work}
\label{sec:Related Work}

%-------------------------------------------------------------------------
\paragraph{Medical Foundation Models}
Foundation models are pre-trained, large-scale models that allow for rapid customization through fine-tuning or in-context learning, as exemplified in~\cite{devlin2018bert,radford2021learning}.
Despite these notable signs of progress, challenges persist in complex tasks like image segmentation, primarily due to the difficulty of obtaining annotated masks. Segment Anything Model (SAM)~\cite{kirillov2023segment}, with extensive training on a dataset of over 1 billion natural images, showcases impressive performance in image segmentation. Particularly, in diverse real-world scenarios, SAM demonstrates powerful capabilities for zero-shot generalization, signifying its potential to address intricate computer vision challenges.

With the surge of Segment Anything Model~\cite{kirillov2023segment}, previous work seeks to apply SAM on medical images~\cite{chai2023ladder,zhang2023input,deng2023sam}. However, empirical studies of SAM's zero-shot capability on medical images~\cite{deng2023segment,hu2023sam,wald2023sam,cheng2023sam} reveal a significant decline when dealing with unseen medical features~\cite{mohapatra2023sam,wang2023sam}. Consequently, recent work explores effective adaptations of prompted SAM on medical datasets. Due to the computational intensity associated with training all parameters of SAM, researchers primarily concentrate on updating a subset of SAM's parameters. Many works require prompts to generalize SAM on medical datasets. For instance, MedSAM~\cite{ma2023segment} curates a large medical image dataset to adapt bound box prompted SAM. Medical SAM Adapter~\cite{wu2023medical} fine-tunes point-prompted SAM using Adaption modules. Several studies transfer SAM from 2D to 3D by adding layers to support volumetric inputs~\cite{gong20233dsam,li2023auto,lei2023medlsam,liu2023samm}. Other prompted SAM adaptations employ exemplar learning~\cite{feng2023cheap} and new prompt mechanisms~\cite{zhang2023self}. 
Apart from prompted SAM adaptations, which require prompts sampled from ground truth during testing,  prompt-free SAM adaptation methods are proposed for medical segmentation where prompts are not necessarily available. AutoSAM~\cite{hu2023efficiently} fine-tunes SAM without prompts by freezing the SAM encoder and adding prediction heads to generate segmentation masks. SAMed~\cite{zhang2023customized} achieves competitive prompt-free results by introducing LoRA~\cite{hu2021lora} layers into the image encoder while using the original decoder. 

This study also introduces a prompt-free version of SAM, designed to enhance efficient finetuning with limited medical data. It differentiates from existing research by introducing a novel hierarchical decoding process that incorporates medical prior knowledge.

\paragraph{Model Fine-Tuning} 
Efficient fine-tuning of foundation models is crucial, with adapters becoming key for integrating new knowledge into pretrained models and providing a more efficient alternative to complete finetuning~\cite{houlsby2019parameter,chen2022vision}. Low Rank Adaptation (LoRA)~\cite{hu2021lora} advocates for a gradual parameter update within transformer blocks, aiming for a low-rank approximation to refine large-scale models. In our proposed H-SAM, we incorporate LoRA adapters into SAM image encoder to avoid over-fitting in medical image dataset adaptation.

\paragraph{Medical Image Segmentation} 
Medical image segmentation refers to partitioning or dividing a medical image into the dense prediction of pixels corresponding to lesions or organs in imaging modalities such as CT~\cite{fu2021review,zhou2019prior,wang2019abdominal,zhou2019semi} and MRI~\cite{zeng2020review,ji2022amos}. With the rapid progress of deep learning, U-Net~\cite{ronneberger2015u} with its elaborate network design ushers a new era for medical image segmentation. 
Building on U-Net's foundation, a series of works emerge to enhance segmentation performance with U-shaped models~\cite{isensee2018nnu,zhou2018unet++,diakogiannis2020resunet}. 
Recent advancements of vision transformers in natural image analysis also prompt exploration in medical image segmentation~\cite{dosovitskiy2020image}. A prevalent network design strategy involves integrating transformer blocks into the U-Net framework, resulting in novel architectures such as TransUnet~\cite{chen2021transunet} and Swin-Unet~\cite{cao2022swin}. MISSFormer~\cite{huang2021missformer} is a u-shaped encoder-decoder network with enhanced transformer blocks. Recent works also explore medical image segmentation in a few-shot setting. RP-Net~\cite{tang2021recurrent} proposed to iteratively capture context relationships using a U-shaped network. CAT-Net and 3D TransUNet~\cite{lin2023few,chen20233d} design a cross masked attention Transformer to focus only on foreground regions between the support image and query image. 

Unlike existing studies, this study is focused on the fine-tuning of large foundational models, specifically SAM, to facilitate more efficient adaptation for medical image segmentation tasks. To achieve this, we introduce a hierarchical decoding strategy designed to optimize SAM's capabilities, thereby unlocking its full potential for efficient and effective fine-tuning on medical tasks.

\section{Methodology}
\label{sec:Methodology}

\subsection{H-SAM Overview}
Given an image $I$ of size $W\times H$, our goal is to predict its corresponding segmentation map of $W\times H$. Each pixel in this map is assigned to a category from a predefined class list, aiming for maximal alignment with the ground truth $gt$.
Our segmentation framework H-SAM is built up upon SAM, integrating a LoRA-adapted image encoder and a simple but effective 2-stage hierarchical decoder. 

\paragraph{LoRA-adapted Image Encoder} As illustrated in Figure~\ref{framework}, H-SAM utilizes the original image encoder of SAM and freezes all layers to preserve pre-learned knowledge. 
Then, we adopt the same LoRA implementation as SAMed~\cite{zhang2023customized} to add a smaller, trainable bypass composed of two low-rank matrices. In line with LoRA, these bypasses first compress the transformer features into a low-rank space. Subsequently, they reproject these condensed features to match the output feature channels of the frozen transformer blocks.  Only these bypass matrices are updated during training, allowing for minor yet effective model adjustments. 
For the prompt encoder, H-SAM does not need any prompt and simply updates a default embedding during training.

\paragraph{Mask Decoder} The original SAM mask decoder consists of a Transformer decoder and a pixel decoder. The Transformer decoder processes image embeddings extracted from the image encoder, employing self-attention mechanisms to evaluate the significance of various image regions and cross-attention mechanisms to focus on relevant areas for segmentation. 
Subsequently, the pixel decoder refines this output, generating a detailed segmentation map, and assigning a class or category to each pixel.

\paragraph{Hierarchical Decoding} Our H-SAM introduces a more intricate two-stage hierarchical decoding procedure. In the first stage, H-SAM employs SAM's original decoder to create a prior (probabilistic) mask, which will be used to guide more intricate decoding in the second stage, as illustrated in Figure~\ref{framework}. 
This second stage mirrors the original one with both a Transformer decoder and a pixel decoder. To enhance image embedding input and optimize cross-attention in the second Transformer decoder, we introduce two novel modules. 
Firstly, a class-balanced, mask-guided self-attention mechanism is proposed to rectify the issue of unbalanced label distribution, thereby enhancing the image embedding for the second-stage Transformer decoder (Sec.~\ref{sec:CMAttn}). Secondly, we incorporate a learnable mask cross-attention mechanism within the second Transformer decoder. This mechanism adeptly modulates the spatial dynamics among various image regions, guided by the information from the prior mask, thereby enhancing the segmentation process (Sec.~\ref{sec:learnMaskAttn}).
Collectively, these decoders constitute a hierarchical Transformer decoder framework.
Further, we propose a hierarchical pixel decoder, inspired by U-Net architecture, to supplement the hierarchical Transformer decoder and further refine the segmentation outcome. Concretely, the pixel decoder in the second stage integrates features from the first-stage pixel decoder through skip connections, enabling the generation of high-resolution predictions (Sec.~\ref{sec:hpixeldec}).

\begin{figure}
  \centering
   \includegraphics[width=1\linewidth]{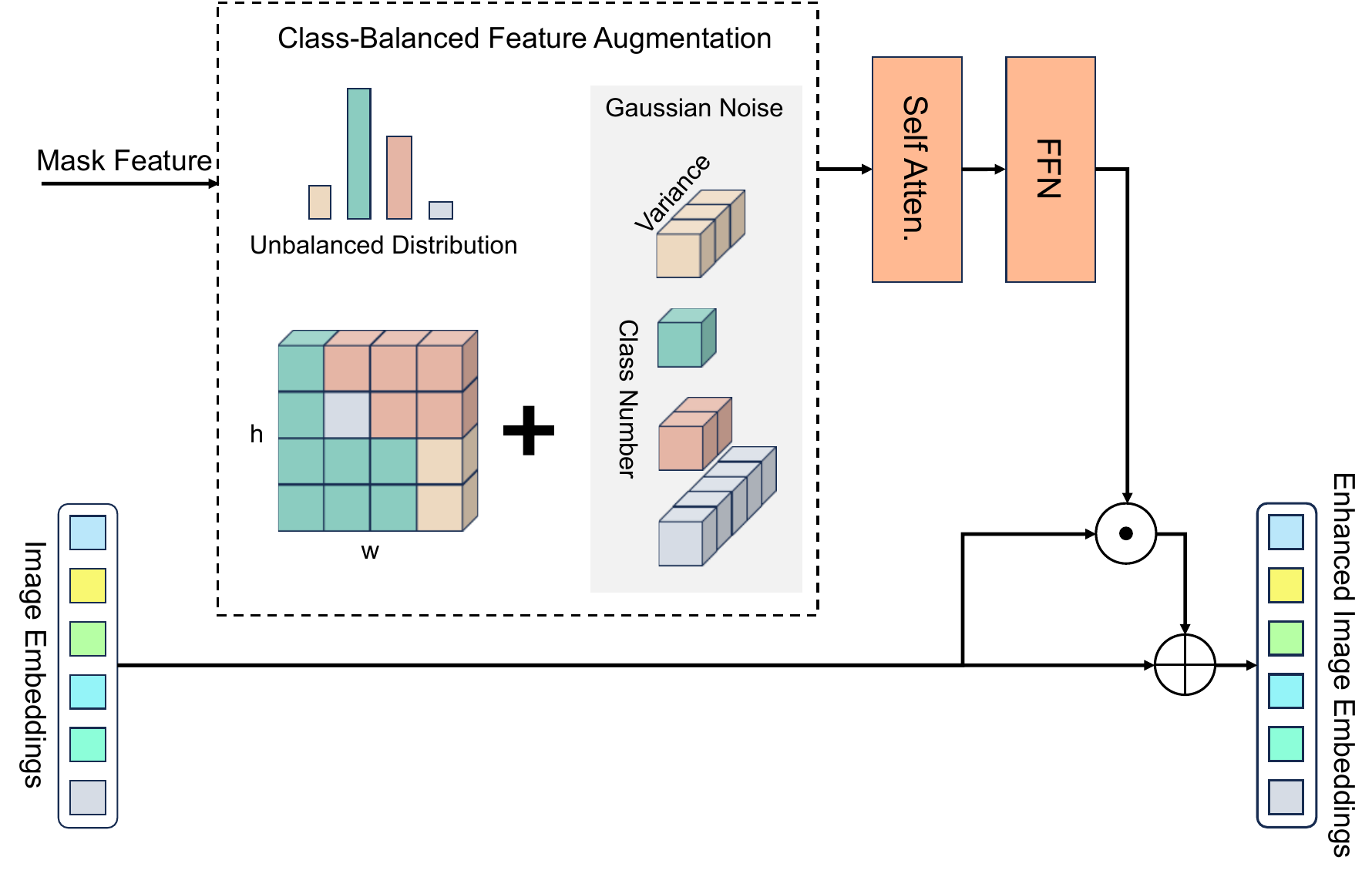}

   \caption{The illustration of Class-Balanced Mask-Guided Self-Attention (CMAttn) block.}
   \label{cmsa}
\end{figure}

\subsection{Enhanced Image Embedding via Class-Balanced Mask-Guided Self-Attention}
\label{sec:CMAttn}

As shown in Figure~\ref{cmsa}, we incorporate a Class-Balanced Mask-Guided Self-Attention (CMAttn) block to enhance the image embeddings as input for the second-stage Transformer decoder. This is particularly helpful in the case where we have an imbalance between the abundant instances in the head categories and the scarcity of instances in the tail categories.
We use a mask feature which is acquired by directly multiplying image embedding without upsampling from the first decoder as the input mask feature for CMAttn. 
Before the self-attention block, we adopt a class-balanced augmentation to introduce more variations to tail categories. Inspired by previous approaches utilizing logit adjustment in long-tail problems \cite{li2022long}, we 
perturb the mask feature with Gaussian noise whose variance is inversely proportional to category sample frequencies:
\begin{equation}
  P(gt\!=\!=i)  +\!\!= \mathcal N(0,var(i)),
  \label{eq:important}
\end{equation}
where $P\in\mathbb{R}^{N\times C\times H\times W}$ is the normalized input mask feature. $\textit{gt}$ is the ground truth mask resized to the same size. $\mathcal N$ is the added Gaussian noise. The variance list is calculated offline and stored as $var$.

After self-attention, we adopt a linear layer to compress the channel dimension, and incorporate the resulting mask feature to input image embeddings using Hadamard product $\odot$. A residual path is designed to retain information from the initial image embedding.

%-------------------------------------------------------------------------
\begin{figure}
  \centering
   \includegraphics[width=.95\linewidth]{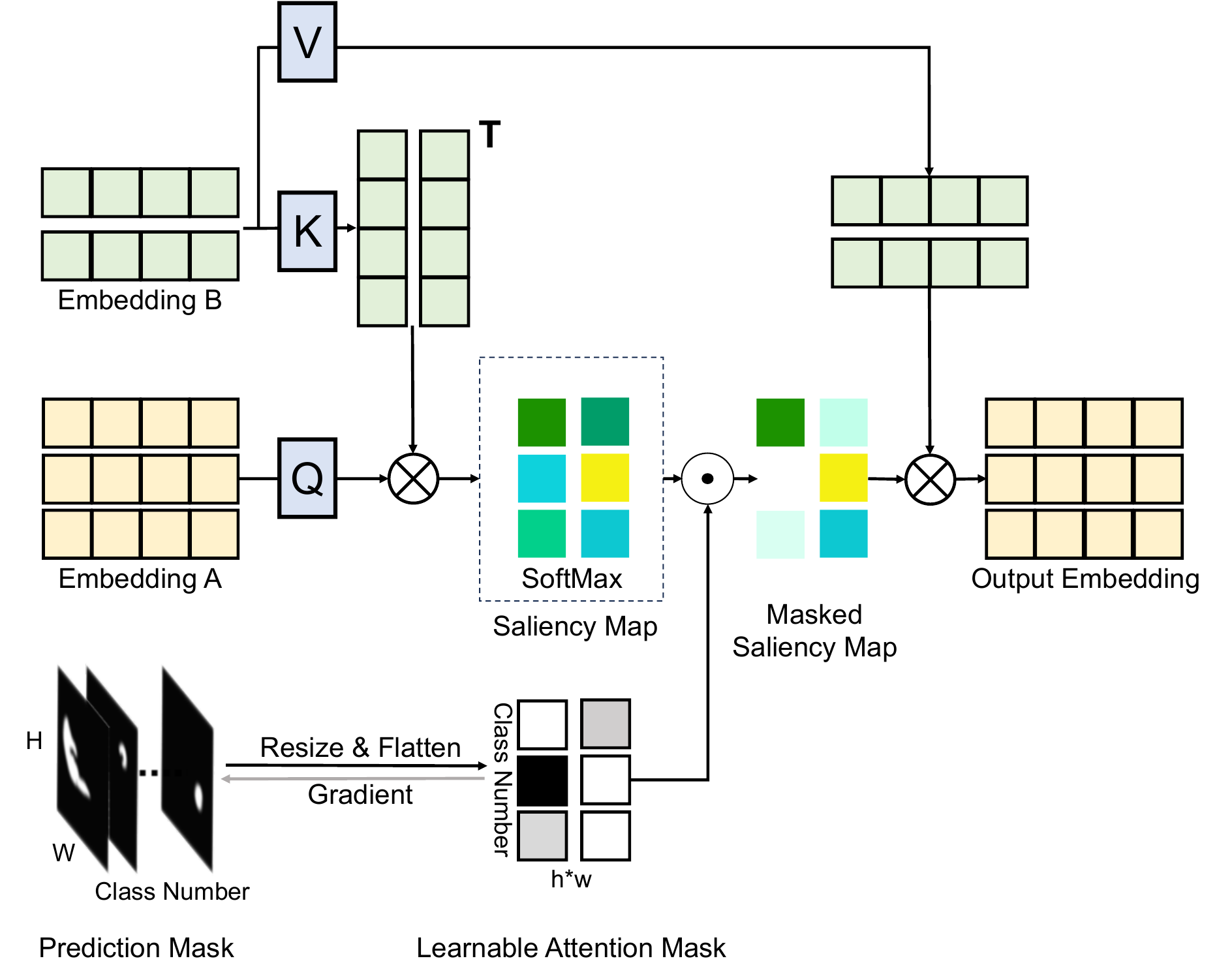}

   \caption{The illustration of learnable mask-attention.}
   \label{mask_attention}
\end{figure}

\subsection{Learnable Mask Cross Attention}
\label{sec:learnMaskAttn}
Mask-attention is a variant of cross-attention first proposed in Mask2Former \cite{cheng2022masked}. Unlike cross-attention which attends to the global context, mask-attention operates only on the area within the predicted mask. Original mask-attention adds a transformed binary mask to the cross attention operation via 
\begin{equation}
  X=softmax(t\textit{}(M)+K Q^T) V+X,
  \label{original_mask_attention}
\end{equation}
where $X$ is the input query feature of the transformer block. 
$K$, $Q$, and $V$ are the key, query and value in cross attention.
$t(M)$ is a function mapping binarized input $\{0,1\}$ to $\{-\infty,0\}$. This mask formulation has two limitations: (1) the gradient of mask $M$ vanishes through $t(M)$; (2) binarized mask $M$ treats all foreground pixels without differentiation, limiting its ability to interpret further information from the mask prior.

To address these limitations, we propose to use learnable mask cross-attention in the second decoding stage as shown in Figure~\ref{mask_attention}, which can be formulated as:
\begin{equation}
  X=M \odot softmax(K Q^T) V+X,
\label{Learnable_mask_attention}
\end{equation}
which employs an untransformed probabilistic map $M$ resized to the same spatial resolution as the saliency map in cross attention.
With element-wise product between the mask and the saliency map, 
masked regions will be ignored by multiplying a near-zero probability. 
This new formulation mitigates the aforementioned limitations and facilitates rapid convergence and better performance. Our learnable mask cross-attention within the second-stage Transformer decoder leverages more information from the probabilistic map and can assign varying degrees of importance to different foreground regions.

%-------------------------------------------------------------------------
\subsection{Hierarchical pixel decoder}
\label{sec:hpixeldec}
Complementing the Transformer decoder, SAM's original pixel decoder directly upsamples the image embedding into a segmentation map of $H/4 \times W/4$.
We argue that this resolution is not capable of capturing some intricate local details and small-scale medical objects in medical image segmentation necessitates the use of U-shaped networks with multiple skip connections.
To enhance the details in the segmentation output, we adopt U-shaped architectures only in pixel decoders with skip connections for effectively handling multi-scale objects in medical images with acceptable computation cost. 
Within our H-SAM hierarchical decoding, the second-stage transformer decoder undergoes meticulously crafted mask-guided designs to propagate the prior mask from the first stage to the next stage (as shown in Sec.~\ref{sec:CMAttn} and Sec.~\ref{sec:learnMaskAttn}). Here we propose the hierarchical pixel decoder designed to complement the enriched object queries derived from the transformer decoder. Similar to the hierarchical Transformer decoder, the hierarchical pixel decoder also consists of two successive pixel decoders, strategically incorporates skip connections to integrate features from the first pixel decoder to the second, and further upsample the resolution from $H/4 \times W/4$ to the full resolution $H \times W$.
Benefiting from skip-connected localized features, the hierarchical pixel decoder will be supplemented with the Transformer decoder to output an enriched representation with enhanced resolution.
 
\paragraph{Training Loss} The training loss combines pixel-wise classification loss and binary mask loss for each segmented prediction:
\begin{align}
\label{eq:matching_criterion}
\mathcal{L} = \lambda_{ce}\mathcal{L}_{ce} + \lambda_{dice}\mathcal{L}_{dice} ,
\end{align}
where the pixel-wise classification loss $\mathcal{L}_{ce}$ and $\mathcal{L}_{dice}$   denote binary cross-entropy loss and dice loss, respectively~\cite{milletari2016v}. For our 2-stage hierarchical structure, there is also a $\lambda_{w}$ for each loss in the 2 stage. The final loss $\mathcal{L}_{total}$ a sum of $\lambda_{w}\mathcal{L}_{stage1}$ and $(1-\lambda_{w})\mathcal{L}_{stage2}$. The parameter $\lambda_{w}$ is set to gradually decrease from 0.8 in a way of exponential decay, at a decay coefficient of 0.005. The first decoder output is supervised by 1/4 resolution ground truth, and the second output by full resolution. The final output is ensembled by taking the average of probabilities from the two outputs.

\paragraph{Deep Supervision} The training loss will be applied to every stage in our hierarchical decoding procedure. The supervisory signals for the stage-1 mask decoder stem from ground truth masks of $H/4 \times W/4$, while the stage-2 mask decoder is directly supervised using original high-resolution ground truth. This ensures thorough supervision and enhances the overall effectiveness of the model.

\section{Experiments}
\label{sec:Experiments}

\begin{table*}
\resizebox{\linewidth}{!}{
  \centering
  \belowrulesep=0pt
  \aboverulesep=0pt
  \begin{tabular}{p{2cm}<{\centering}|c|cccccccc|cc}
    \toprule
     \makecell{Training set} & \makecell{Method}&{Spleen}&\makecell{Right\\Kidney}&\makecell{Left\\Kidney}&Gallbladder&Liver&Stomach&Aorta&Pancreas& \makecell{Mean Dice$\uparrow$\\(\%)}  &HD $\downarrow$\\
    \midrule
    \multirow{4}{*}{10$\%$} & AutoSAM~\cite{hu2023efficiently} &68.80&77.44&76.53&24.87&88.06&52.70&75.19&34.58
     &55.69&31.67\\
     & SAM Adapter~\cite{chen2023sam}  &72.42&68.38&66.77&22.38&89.69&53.15&66.74&26.76
     &58.28&54.22\\
     & SAMed~\cite{zhang2023customized} &85.82&82.25&82.62&63.15&92.72&67.20&78.72&52.12
     &75.57&23.02\\
      & H-SAM (ours) &90.21&84.16&85.65&70.70&94.29&76.10&85.54&56.17
     &\textbf{80.35}&\textbf{15.54}\\
     \cline{1-12}
     \multirow{9}{*}{\makecell{Fully\\Supervised}} &  TransUnet~\cite{chen2021transunet} &87.23& 63.13&81.87&77.02&94.08&55.86&85.08& 75.62&77.48&31.69\\
    & SwinUnet~\cite{cao2022swin} &85.47&66.53&83.28&79.61& 94.29&56.58&90.66&76.60
&79.13&21.55\\
     & TransDeepLab~\cite{azad2022transdeeplab} &86.04& 69.16&84.08&79.88&93.53&61.19&89.00&78.40&80.16&21.25\\
      & DAE-Former~\cite{azad2023dae}&88.96&72.30&86.08&80.88&94.98&65.12&91.94&79.19&82.43&17.46\\
      & MERIT~\cite{rahman2023multi} &92.01&84.85&87.79&74.40&95.26&85.38&87.71&71.81&84.90& 13.22\\
      \cline{2-12}
      & AutoSAM~\cite{hu2023efficiently} &80.54&80.02&79.60& 41.37&89.24&61.14&82.56&44.22&62.08&27.56\\
      & SAM Adapter~\cite{chen2023sam} &83.68&79.00&79.02&57.49&92.67&69.48&77.93&43.07&72.80&33.08\\
      & SAMed~\cite{zhang2023customized} &87.77&69.11&80.45& 79.95&94.80&72.17&88.72&82.06&81.88&20.64\\
      & \textbf{H-SAM (ours)} &93.34(8.22)&89.93(9.59)&91.88(8.25)&73.49(23.01)&95.72(1.49)&87.10(7.14)&89.38(2.82)&71.11(10.27)&\textbf{86.49}&\textbf{8.18}\\
    \bottomrule
  \end{tabular}
  }
  \vspace{-0.5em}
  \caption{Comparison to state-of-the-art models on Synapse multi-organ CT dataset with both few-shot and fully-supervised settings. Our model shows outstanding results in both of the training settings. The value in ($\cdot$) is standard deviation.}
  \label{tab:synapse}
\end{table*}
%-------------------------------------------------------------------------
\subsection{Dataset and Evaluation}

We conduct experiments of multi-organ semantic segmentation on three medical datasets, including Synapse multi-organ CT~\cite{landman2015miccai}, the left atrial (LA) dataset~\cite{chen2019multi}, and PROMISE12~\cite{litjens2014evaluation}. We utilize Dice coefficient and the average Hausdorff distance (HD) as evaluation metrics.

\paragraph{Synapse Multi-Organ CT.} The Synapse dataset is from MICCAI 2015 Multi-Atlas Abdomen Labeling Challenge, which contains 3779 axial contrast enhanced abdominal CT images in total. The training set contains 2212 axial slices. We strictly follow TransUnet~\cite{chen2021transunet} and SAMed~\cite{zhang2023customized} for dataset split and data preprocessing. The dataset is split into 18 training cases and 12 test cases. The CT volumes for Synapse dataset of each volume contain 85 to 198 slices. The resolution of Synapse dataset is 512$\times$512 during few-shot training, and 224$\times$224 during fully-supervised training. We evaluate eight abdominal organs (aorta, gallbladder, spleen, left kidney, right kidney, liver, pancreas, stomach), following TransUnet~\cite{chen2021transunet}.

\paragraph{LA.} The left atrial (LA) dataset is from 2018 Atrial Segmentation Challenge~\cite{chen2019multi}. We strictly follow UA-MT~\cite{yu2019uncertainty} and BCP~\cite{bai2023bidirectional} for data split and data preprocess. Specifically, LA dataset is split into 80 scans for training and 20 scans for evaluation. And we keep 4(5\%)  scans as labeled data while the rest scans in the training set are treated as unlabeled data for these two settings respectively. And we resize each 2D slice into 512$\times$512 during training. Note that we do not use the unlabeled data for training H-SAM. Instead, we only use these selected 4 labeled scans to train our model.

\paragraph{PROMISE12.} PROMISE2012 dataset is from the Prostate MR Image Segmentation 2012~\cite{litjens2014evaluation}. It contains 50 3D
transversal T2-weighted MR images of the prostate
with manual binary prostate gland segmentation and
is obtained from multiple centers with different acquisition protocols. During the experiments, we strictly follow MLB-Seg~\cite{wei2023consistency} for data split and data preprocessing. Specifically, we split it into
40 / 10 cases for training / evaluation. 3 out of 40 are selected as data with labels while the rest 37 scans are used as unlabeled data. The resolution of PROMISE12 dataset is 512$\times$512. Again, note that only the 3 labeled cases are used for training H-SAM.

%-------------------------------------------------------------------------
\subsection{Implementation details}
All our implementation is in PyTorch and we train all our models on 4 NVIDIA RTX A5000 GPUs. During training, we adopt a data augmentation combination of elastic deformation, rotation, and scaling. The training loss is a combination of Cross-Entropy loss and Dice loss. We adopt the same LoRA settings as SAMed, in which the rank of LoRA is set to 4. A ViT-B and a ViT-L backbone are adopted separately for few-shot and fully-supervised training. For a fair comparison, we utilize the same resolution of 224$\times$224 for fully supervised training on Synapse as other SAM variants and SOTA methods. The maximal training epoch is set to 300. The optimizer algorithm used for updating is based on the AdamW, and $\beta$1, $\beta$2, and weight decay are set to 0.9, 0.999, and 0.1. 

%-------------------------------------------------------------------------

\subsection{Results}
\paragraph{Synapse Multi-Organ CT}
In Table~\ref{tab:synapse}, H-SAM shows outstanding few-shot transferability with limited seen medical images (10$\%$). Compared to other SAM prompt-free variants: Auto SAM~\cite{hu2023efficiently}, SAM Adapter~\cite{chen2023sam} and SAMed~\cite{zhang2023customized}, our H-SAM reports results of 80.35\% with limited training scans equal to only 1 volume, which outperforms other SAM adaptation variants by a large margin ($\approx$5\%). We also evaluate our H-SAM under a fully-supervised setting on Synapse multi-organ CT dataset. We evaluate our model in a fair comparison with the prompt-free setting with several state-of-the-art methods, including TransUnet~\cite{chen2021transunet}, SwinUnet~\cite{cao2022swin}, TransDeepLab~\cite{azad2022transdeeplab}, DAE-Former~\cite{azad2023dae} and MERIT\cite{rahman2023multi}, along with other SAM prompt-free variants: Auto SAM~\cite{hu2023efficiently}, SAM Adapter~\cite{chen2023sam} and SAMed~\cite{zhang2023customized}. Our method achieves promising results in multi-organ segmentation with an 86.49\% Mean Dice Coefficient, which is higher than newly released medical segmentation networks DAE-Former (82.43\%) and MERIT (84.90\%). H-SAM also easily outperforms other prompt-free SAM variants (86.24\% vs. 81.88\%).

\begin{table}[t!]
\centering
\resizebox{\linewidth}{!}{
  \centering
  \belowrulesep=0pt
  \aboverulesep=0pt
  \begin{tabular}{c|cc|c}
    \toprule
     \multirow{2}{*}{Methods} &\multicolumn{2}{c|}{Scans used} &\multirow{2}{*}{Mean Dice (\%)$\uparrow$} \\
     &Labeled & Unlabeled &\\
    \midrule
    UA-MT~\cite{yu2019uncertainty}&\multirow{7}{*}{4(5\%)} &\multirow{7}{*}{76(95\%)} &82.26\\
    SASSNet~\cite{li2020shape}& & &81.60\\
    DTC~\cite{luo2021semi}& & &81.25\\
    URPC~\cite{luo2021efficient} & & &82.48 \\
    MC-Net~\cite{wu2021semi} & & &83.59\\
    SS-Net~\cite{wu2022exploring}& & &86.33\\
    BCP~\cite{bai2023bidirectional}& & &88.02\\ \hline
    nnUnet~\cite{isensee2021nnu} &\multirow{5}{*}{4(5\%)} &\multirow{5}{*}{0(0\%)}&64.02\\
    AutoSAM~\cite{hu2023efficiently} &&&74.73\\
    SAM Adapter~\cite{zhang2023customized}& & &82.79\\
    SAMed~\cite{zhang2023customized}& & &87.72\\
    Ours& & & \textbf{89.22}\\
    \bottomrule
  \end{tabular}
  }
  \caption{Results of LA dataset under semi-supervision using 4 labeled scans.}
  \label{tab:LA semi results_4}
  \vspace{-.5em}
\end{table}

\paragraph{LA}
As shown in Table \ref{tab:LA semi results_4}, we conduct the few-shot semantic segmentation experiment on LA dataset. Here, we present results using 4 labeled scans (5\% of the dataset) for training. 
We compare our method with two categories of approaches: 1) SAM efficient adaptation methods including AutoSAM~\cite{hu2023efficiently}, SAM Adapter, and SAMed~\cite{zhang2023customized} and 2) semi-supervised methods such as UA-MT~\cite{yu2019uncertainty}, MC-Net~\cite{wu2021semi}, SS-Net~\cite{wu2022exploring}, and BCP~\cite{bai2023bidirectional}. We also compare our results with nnUnet~\cite{isensee2021nnu}, a well-known baseline for medical image segmentation. To ensure a fair comparison, we adhered to the data split protocols established in UA-MT~\cite{yu2019uncertainty} and BCP~\cite{bai2023bidirectional}, employing identical sets of 4 labeled scans for our few-shot experiments. The semi-supervised methods also utilized the remaining 76 unlabeled scans during their training. 
And all the methods were evaluated on the same test dataset. Our proposed H-SAM method outperforms both the SAM efficient adaptation and semi-supervised methods under both split settings. This underscores the ability of H-SAM to achieve superior or comparable results with significantly less data compared to semi-supervised methods. 
Under the few-shot setting, H-SAM also demonstrates superior improvements in dice coefficients compared to SAMed~\cite{zhang2023customized} (89.22\% vs. 87.72\%), indicating the effectiveness of H-SAM in few-shot learning. 

\paragraph{PROMISE12} We also conducted the few-shot semantic segmentation experiment on PROMISE12 dataset. Similar to the LA dataset, we compared H-SAM against SAM efficient adaptation methods (AutoSAM~\cite{hu2023efficiently}, SAM Adapter~\cite{chen2023sam}, and SAMed~\cite{zhang2023customized}), nnUnet~\cite{isensee2021nnu}, and semi-supervised methods (UA-MT~\cite{yu2019uncertainty}, MC-Net~\cite{wu2021semi}, SS-Net~\cite{wu2022exploring}, and MLB-Seg~\cite{wei2023consistency}).
For a fair comparison with the semi-supervised methods, we strictly follow MLB-Seg~\cite{wei2023consistency} and use the same 3 labeled cases as our training set under the few-shot setting, while the semi-supervised methods incorporated an additional 37 unlabeled cases. And all the methods are tested on the same test dataset. As shown in Table~\ref{tab:PROMISE12 semi results}, H-SAM achieved a significant improvement in Dice coefficients (approximately 10.94\%) over the semi-supervised method MLB-Seg, despite being trained on only three labeled cases.  This result highlights the efficiency of H-SAM in leveraging limited labeled data. Compared to the SAM efficient adaptation method SAMed~\cite{zhang2023customized}, our method also demonstrated superior performance (87.27\% vs. 86.00\%), further establishing its effectiveness in few-shot semantic segmentation tasks.

\begin{table}
\centering
\resizebox{\linewidth}{!}{
  \centering
  \belowrulesep=0pt
  \aboverulesep=0pt
  \begin{tabular}{c|cc|c}
    \toprule
     \multirow{2}{*}{Methods} &\multicolumn{2}{c|}{Scans used} &\multirow{2}{*}{Mean Dice (\%)$\uparrow$} \\
     &Labeled & Unlabeled &\\
    \midrule
    UA-MT~\cite{yu2019uncertainty} &\multirow{7}{*}{3(7.5\%)} &\multirow{7}{*}{37(92.5\%)}& 65.05 \\
    DTC~\cite{luo2021semi}& & &63.44\\
    SASSNet~\cite{li2020shape}& & &73.43\\
    MC-Net~\cite{wu2021semi} & & &72.66\\
    SS-Net~\cite{wu2022exploring}& & &73.19\\
    Self-Paced~\cite{peng2021self} & & & 74.02 \\
    MLB-Seg~\cite{wei2023consistency}& & &78.27\\\hline
    nnUnet~\cite{isensee2021nnu} &\multirow{5}{*}{3(7.5\%)} &\multirow{5}{*}{0(0\%)} &84.22\\
    AutoSAM~\cite{hu2023efficiently}&&&68.40\\
    SAM Adapter~\cite{chen2023sam}& & &75.45\\
    SAMed~\cite{zhang2023customized}& & &86.00\\
    Ours& & &\textbf{87.27}\\
    \bottomrule
  \end{tabular}
  }
  \caption{Results of PROMISE12 dataset under semi-supervision.}
  \label{tab:PROMISE12 semi results}
\end{table}

\begin{table}
\centering
\resizebox{\linewidth}{!}{
  \centering
  \belowrulesep=0pt
  \aboverulesep=0pt
  \begin{tabular}{ccc|cccccccc|c}
    \toprule
    \makecell*[c]{Learnable\\Mask-Attention} & \makecell*[c]{Hierarchical\\Pixel Decoder} & \makecell*[c]{CM\\Self-Attention}& Mean Dice (\%)\\
    \midrule
     \XSolidBrush & \XSolidBrush & \XSolidBrush &75.57 \\
     \Checkmark & \XSolidBrush & \XSolidBrush &77.68\\
     \Checkmark & \Checkmark & \XSolidBrush &78.58\\
     \XSolidBrush & \Checkmark & \XSolidBrush &77.05\\
     \XSolidBrush & \Checkmark & \Checkmark &79.03\\
     \XSolidBrush & \XSolidBrush & \Checkmark &77.71\\
     \Checkmark & \XSolidBrush & \Checkmark &78.76 \\
     \Checkmark & \Checkmark & \Checkmark&\textbf{80.35}\\
    
    \bottomrule
  \end{tabular}
  }
  \caption{Effectiveness of the key contributions in H-SAM: Learnable Mask-Attention, CMAttn and Hierarchical Pixel Decoder.}
  \label{tab:ablation_whole}
\end{table}

\begin{table}[t!]
\centering
\resizebox{0.8\linewidth}{!}{
  \centering
  \belowrulesep=0pt
  \aboverulesep=0pt
  \begin{tabular}{c|c}
    \toprule
     Methods& \makecell{Mean Dice (\%)} \\
    \midrule
    w/o. mask-attention &75.57\\
    original mask-attention&75.61\\
    learnable mask-attention (ours)&\textbf{77.68}\\
    \bottomrule
  \end{tabular}
  }
  \caption{Effectiveness of our learnable mask-attention against original mask-attention and baseline. }
  \label{tab:mask_attention}
\end{table}

\subsection{Ablation study}
\paragraph{Effectiveness of Learnable Mask Cross Attention}
 As shown in Table~\ref{tab:ablation_whole}, our learnable mask attention improves the baseline by 2.1\%. Within the hierarchical decoding structure of H-SAM, our learnable mask cross attention shows no signal of performance saturation with other two key contributions: CMAttn and Hierarchical Pixel Decoder. We also compare learnable mask cross attention against normal cross attention and unlearnable mask attention in Table~\ref{tab:mask_attention}. Note that, unlearnable mask attention operation brings little improvement due to the lack of gradient backpropagation. Inversely, our proposed learnable mask cross attention brings an immediate performance increase of 2.1\% with jointing training and inheritance of mask-guided prior.

\paragraph{Effectiveness of CMAttn}
 In Table~\ref{tab:ablation_whole}, we present the ablation of Class-Balanced Mask-Guided Self-Attention (CMAttn). The CMAttn alone brings a 1.2\% improvement to the baseline, showing that SAM benefits from a more informed image embedding as input for the mask decoder. Combined with Learnable Mask-Attention, the two mask-guided implementations improve the baseline model by 3.2\% in terms of Mean Dice. 

\paragraph{Effectiveness of Hierarchical Pixel Decoder}
 In Table~\ref{tab:ablation_whole}, the hierarchical pixel decoder promotes the dice coefficient by 2.2\%. Then we evaluate the effectiveness of a combination of mask-guided implementations with the hierarchical pixel decoder by adding or disabling one of them each time. Table~\ref{tab:ablation_whole} shows an improvement of 3.0\% and 3.5\%, respectively combined with Learnable Mask-Attention and CMAttn.

\subsection{Efficiency Analysis}
To show that the advantages of our hierarchical decoding are not achieved by additional training cost in the decoding section, here we conduct ablation experiments between our model and other prompt-free SAM variants in terms of total parameters and performance. AutoSAM~\cite{hu2023efficiently} is not listed in the table because it freezes the image encoder with no adapters. The original lightweight SAM mask decoder consists of only 2 transformer layers. SAMed~\cite{zhang2023customized} adds LoRA adapter in the image encoder, while remaining a default SAM mask decoder with 2 transformer layers. SAM Adapter~\cite{chen2023sam} injects adapter layers in both the image encoder and mask decoder. In general, the training cost of our H-SAM is equal to SAMed with 1 additional lightweight mask decoder. For a fair comparison, we double and dribble the number of SAMed transformer layers into 4 and 6, to achieve a comparable and even larger parameter scale than the training cost in our H-SAM. As shown in Table~\ref{tab:parameter}, SAMed~\cite{zhang2023customized} benefits from increased transformer layers. However, the performance promotion comes with huge computation costs. H-SAM consistently outperforms SAMed. Compared with SAM Adapter, H-SAM shows a 7.5\% performance improvement with a 20M lesser parameter scale. The results prove the superiority of our H-SAM in wisely generating finer medical segmentation with little extra computation cost.

\begin{table}
\resizebox{\linewidth}{!}{
  \centering
  \belowrulesep=0pt
  \aboverulesep=0pt
  \begin{tabular}{c|c|c|c}
    \toprule
     Methods&\makecell{Transformer\\Layers}&\makecell{Total\\Parameter}&Mean Dice (\%)\\
    \midrule
    \multirow{3}{*}{SAMed~\cite{zhang2023customized}} &2&108.8M&75.57\\
    &4&112.5M&76.80\\
     &6&116.2M&78.05\\
     \hline
     SAM Adapter~\cite{chen2023sam}&2&131.5M&72.80\\
     \hline
     H-SAM(ours)&4&112.3M&\textbf{80.35}\\
    \bottomrule
  \end{tabular}
  }
  \caption{Efficiency analysis of our H-SAM against deeper default SAM mask decoder. H-SAM shows better performance with fewer parameters.}
  \label{tab:parameter}
\end{table}

\begin{figure}
  \centering
   \includegraphics[width=\linewidth]{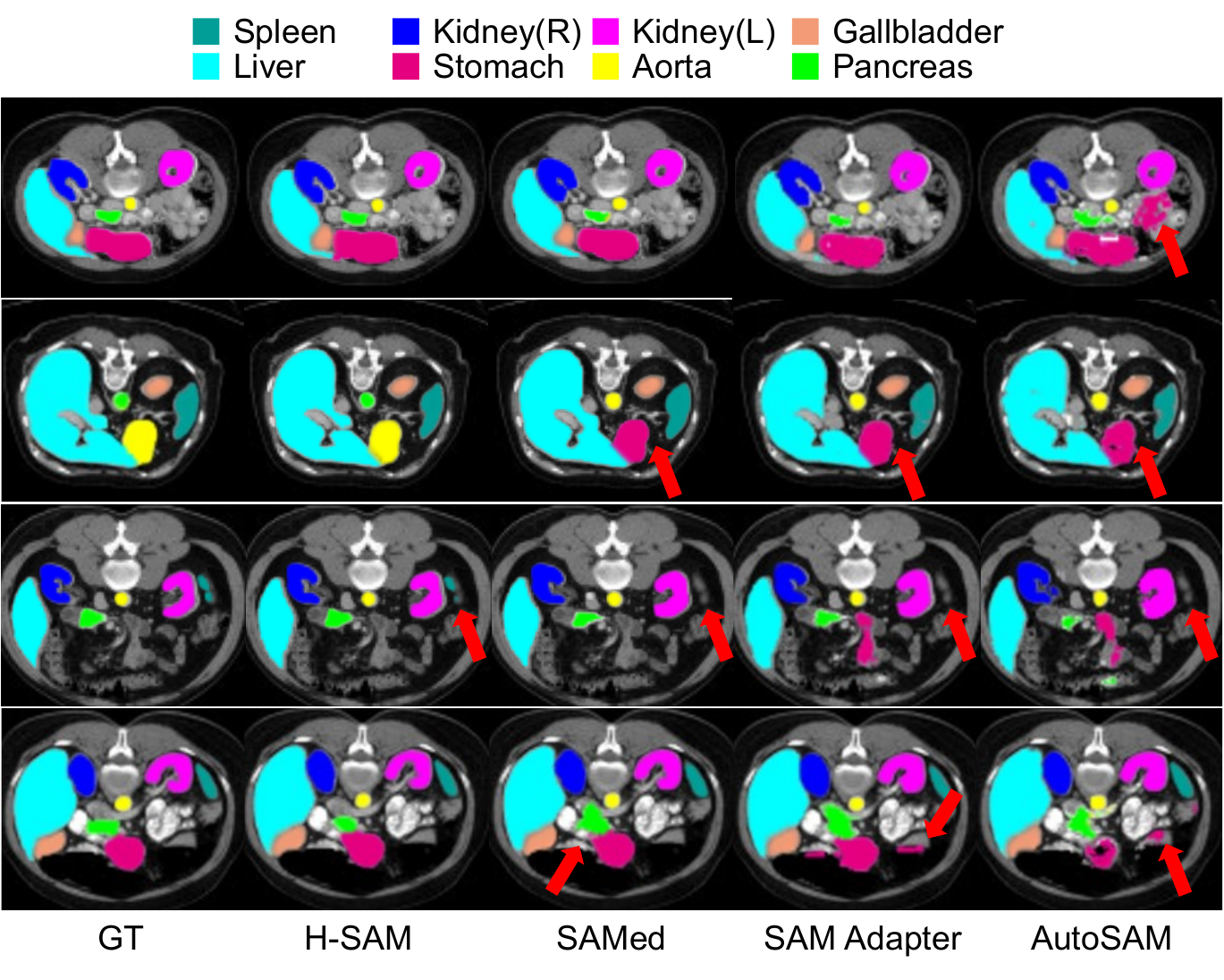}

   \caption{The qualitative results of H-SAM and other SAM variants, including SAMed, SAM Adapter, and AutoSAM.}
   \label{qualitative}
\end{figure}
%-------------------------------------------------------------------------
\subsection{Qualitative Results}
As shown in Figure~\ref{qualitative}, we compare our H-SAM to other prompt-free medical SAM variants, including AutoSAM~\cite{chen2023sam}, SAM Adapter~\cite{hu2023efficiently} and SAMed~\cite{zhang2023customized}. As pointed out in the first row and the last row, compared to other SAM variants, H-SAM provides a precise mask prediction with lesser noise. In the second row, where other methods misrecognize Aorta and Pancreas to Stomach and Aorta, H-SAM provides correctly attributes each organ to their categories. H-SAM also performs superiorly with small-scale organs. In the third row, while all other variants miss Spleen, only H-SAM provides correct prediction for all of the organs.
%-------------------------------------------------------------------------
\section{Conclusion}
\label{sec:Conclusion}

We present H-SAM, which is a simple and efficient hierarchical mask decoder for adaptation of Segment Anything Model on medical image segmentation. Using a probabilistic map from a default decoder as prior to guide finer medical segmentation in the sequential decoding unit, our H-SAM puts forward a new direction of SAM adaptation. Notably, H-SAM achieves this superior performance without relying on any unlabeled data, surpassing even state-of-the-art semi-supervised models that use extensive unlabeled datasets in various medical imaging contexts. This underscores H-SAM's significant potential in advancing the field of medical image segmentation, offering a robust, efficient, and data-economic solution.

{
    \small
    \bibliographystyle{ieeenat_fullname}
    \bibliography{main}
}
\clearpage
\setcounter{page}{1}
\maketitlesupplementary
\begin{appendices}
In this supplementary material we first provide more implementation details as in \cref{sec:Methodology} and \cref{sec:Experiments} for training strategies and datasets. Then, we present more visualized results with zoom-in analysis for H-SAM. Finally, we present an organ-by-organ analysis of the result of H-SAM on Synapse CT dataset.

\section{Implementation Details}
\subsection{Training strategy}
We provide training strategy and hyper-parameter setting as supplementary for \cref{sec:Methodology} and \cref{sec:Experiments}. We adopt warmup during training. As shown in Table~\ref{tab:training}, the initial learning rate is set to 0.0025, and the warmup period is set to 250. The training loss we use is a combination of Dice loss $\mathcal{L}_{dice}$ and MSE loss $\mathcal{L}_{ce}$. To be specific, as shown in Table~\ref{tab:2_stage_setting}, the weight of each loss is 0.9 for $\mathcal{L}_{dice}$ and 0.1 for $\mathcal{L}_{ce}$. For our 2-stage hierarchical structure, there is also a $\lambda_{w}$ for each loss in the 2 stages. The final loss $\mathcal{L}_{total}$ a sum of $\lambda_{w}\mathcal{L}_{stage1}$ and $(1-\lambda_{w})\mathcal{L}_{stage2}$. The parameter $\lambda_{w}$ is set to gradually decrease in the way of exponential decay, from 0.4 to 0 in 300 training epochs. The first decoder output is supervised by 1/4 resolution ground truth, and the second output by full resolution. The final output is ensembled from the two outputs, where we utilize a mean value of results from the two stages.

\subsection{Additional datasets information.}
In \cref{sec:Experiments}, we present our dataset settings. The Synapse dataset we experiment on is from MICCAI 2015 Multi-Atlas Abdomen Labeling Challenge, containing 3779 axial contrast-enhanced abdominal CT images in total and the training set contains 2212 axial slices. We follow TransUnet to evaluate eight abdominal organs (aorta, gallbladder, spleen, left kidney, right kidney, liver, pancreas, stomach). For fully-supervised training, we strictly follow TransUnet of the division between training and testing cases. For the few-shot setting on the Synapse dataset, we adopt a slice-based dataset selection. We select 10\% training data, i.e., 221 slices, \emph{randomly} from different subjects in the training volumes, which contains 2212 axial slices in total. 

\begin{table}
\centering
\resizebox{0.6\linewidth}{!}{
  \centering
  \belowrulesep=0pt
  \aboverulesep=0pt
  \begin{tabular}{c|c}
    \toprule
    Config&Setting\\
    \midrule
     Optimizer&AdamW\\
     Learning rate&2.5e-3\\
     Batch size&32\\
     Weight decay&0.1\\
     Optimizer momentum&\makecell{$\beta_1=0.9$\\ $\beta_2=0.999$}\\
     Warmup period&250\\
    \bottomrule
  \end{tabular}
  }
  \caption{Training setting}
  \label{tab:training}
\end{table}

\begin{table}
\centering
\resizebox{0.5\linewidth}{!}{
  \centering
  \belowrulesep=0pt
  \aboverulesep=0pt
  \begin{tabular}{c|c}
    \toprule
    \makecell{Hyper parameter}&Setting\\
    \midrule
     $\mathcal{L}_{dice}$&0.9\\
     $\mathcal{L}_{ce}$&0.1\\
     $\lambda_{w_{start}}$&0.4\\
     $\lambda_{w_{end}}$&0\\
    \bottomrule
  \end{tabular}
  }
  \caption{2 stage hyper-parameter setting}
  \label{tab:2_stage_setting}
\end{table}

\section{Additional ablation analysis}
\subsection{Additional organ-by-organ analysis}
In Table~\ref{tab:ablation_whole}, we provide an ablation study of the Effectiveness of the key contributions in H-SAM: Learnable Mask-Attention, CMAttn, and Hierarchical Pixel Decoder. In Table~\ref{tab:supp_ablation}, here we present an additional organ-by-organ analysis to further prove the validity of H-SAM's innovation. The implementation of Learnable Mask-Attention alone brings a 2.1\% improvement in terms of mean dice. Learnable Mask-Attention also achieves the highest 94.43\% results for the organ Liver. Both its combination with CMAttn and Hierarchical Pixel Decoder achieves promising results on some relatively small-scale organs, such Pancreas (58.18\%) and Aorta (84.37\%). The combination of all three implementations shows promising results on most organs, reflecting the meticulous design of our hierarchical decoding strategy.

\subsection{Additional analysis under Synapse semi-supervised setting}
For the few-shot setting on the Synapse dataset, our H-SAM shows outstanding performance under a slice-based dataset selection. We also validate H-SAM in volume-based dataset selection strictly following the same training split (5 subjects) in MagicNet. As shown in Table~\ref{tab:synapse semi results}, similar to our observation for PROMISE12 and LA, H-SAM also outperforms the SOTA semi-supervised method MagicNet~\cite{chen2023magicnet} without using any unlabeled data.

\begin{table}[H]
\scriptsize
\centering
\resizebox{1\linewidth}{!}{
  \centering
  \belowrulesep=0pt
  \aboverulesep=0pt
  \begin{tabular}{c|cc|c}
  \toprule
     \multirow{2}{*}{Methods} &\multicolumn{2}{c|}{Scans used} &\multirow{2}{*}{Mean Dice (\%)$\uparrow$} \\
     &Labeled & Unlabeled &\\
    \midrule
    SS-Net~[61]&\multirow{3}{*}{5(30\%)} & \multirow{3}{*}{13(70\%)}& 56.74\\
    UA-MT~[62] &&&61.20\\
    MagicNet~[R1]&&&75.53\\\hline
    H-SAM (ours)&5(30\%)&0(0\%)&\textbf{79.36}\\
    \bottomrule
  \end{tabular}
  }
  % \vspace{-0.4cm}
  \caption{Semi-supervised results on Synapse Dataset}
  \label{tab:synapse semi results}
\end{table}

\begin{table}[H]
\centering
\resizebox{0.7\linewidth}{!}{
\centering
  \belowrulesep=0pt
  \aboverulesep=0pt
  \begin{tabular}{c|c|c}
    \toprule
     Rank size&Mean Dice (\%)&Mean HD\\
    \midrule
    1&71.14&27.03\\
    4&80.35&15.54\\
    8&79.15&16.19\\
    16&76.14&16.30\\
    
    \bottomrule
  \end{tabular}
  }
  \caption{Ablation study on rank size of LoRA layers}
  \label{tab:lora}
\end{table}

\subsection{Ablation study on the LoRA component}
 In Table~\ref{tab:lora}, we discuss the effectiveness of the layers of LoRA component. We discover the performance of H-SAM increases to rank=4, but the performance drops gradually when the rank is too large.

\begin{table*}
\resizebox{\linewidth}{!}{
  \centering
  \belowrulesep=0pt
  \aboverulesep=0pt
  \begin{tabular}{ccc|cccccccc|c}
    \toprule
    \makecell*[c]{Learnable\\Mask-Attention} & \makecell*[c]{Hierarchical\\Pixel Decoder} & \makecell*[c]{CM\\Self-Attention}&{Spleen}&\makecell{Right\\Kidney}&\makecell{Left\\Kidney}&Gallbladder&Liver&Stomach&Aorta&Pancreas& Mean Dice (\%)\\
    \midrule
     \XSolidBrush & \XSolidBrush & \XSolidBrush &85.82&82.26&82.62&63.15&92.71&67.20&78.72&52.12&75.57 \\
     \Checkmark & \XSolidBrush & \XSolidBrush &89.56&84.18&82.06&62.48&\textbf{94.43}&70.97&85.22&52.58&77.68\\
     \Checkmark & \Checkmark & \XSolidBrush &89.91&83.93&79.70&\textbf{70.87}&94.16&70.85&82.47&56.72&78.58\\
     \XSolidBrush & \Checkmark & \XSolidBrush &87.32&84.78&79.85&69.39&93.86&68.48&79.43&53.31&77.05\\
     \XSolidBrush & \Checkmark & \Checkmark &89.11&\textbf{85.04}&83.77&69.79&94.00&\textbf{77.93}&81.71&50.94&79.03\\
     \XSolidBrush & \XSolidBrush & \Checkmark &89.86&84.08&82.38&65.02&94.05&73.82&81.87&50.61&77.71\\
     \Checkmark & \XSolidBrush & \Checkmark &87.51&83.98&80.95&65.25&94.13&75.66&84.37&\textbf{58.18}&78.76 \\
     \Checkmark & \Checkmark & \Checkmark&\textbf{90.21}&84.16&\textbf{85.65}&70.70&94.29&76.10&\textbf{85.54}&56.17&\textbf{80.35}\\
    
    \bottomrule
  \end{tabular}
  }
  \caption{Additional ablation result of Learnable Mask-Attention, CMAttn, and Hierarchical Pixel Decoder.}
  \label{tab:supp_ablation}
\end{table*}

\section{Visualization}

\subsection{Zoom-in analysis}
\begin{figure}[H]
  \centering
   \includegraphics[width=0.8\linewidth]{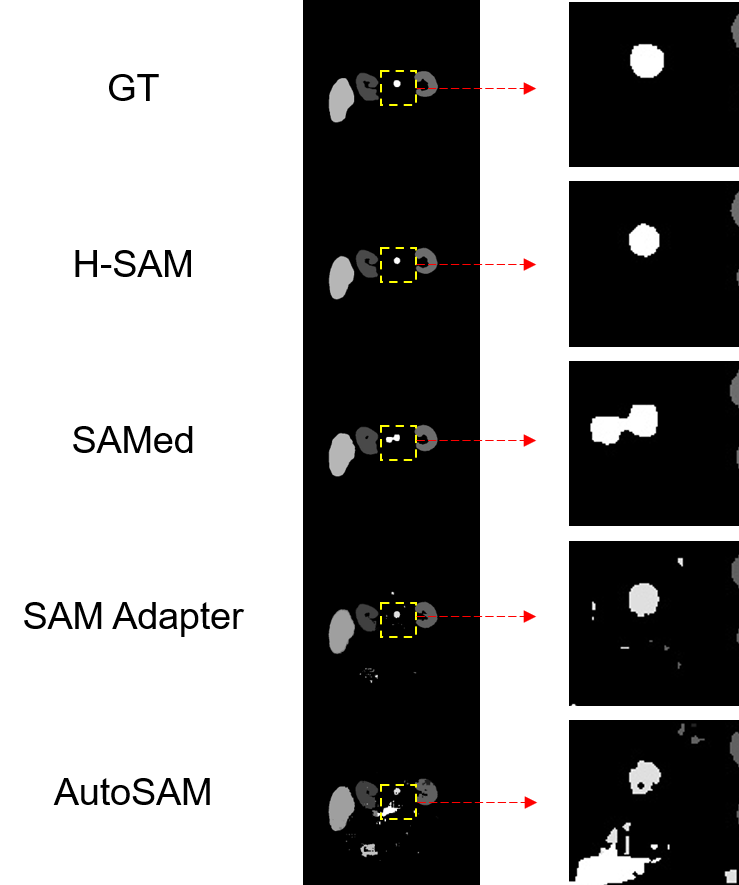}

   \caption{The zoom-in analysis of H-SAM results against other SAM prompt-free variants. H-SAM performs precise segmentation for small-scale organs.}
   \label{zoom}
\end{figure}

As shown in Figure~\ref{zoom} is the zoom-in visualization of H-SAM results against other SAM prompt-free variants. On the Synapse multi-organ CT dataset, H-SAM performs precise segmentation for small-scale organs. The pancreas marked as yellow in the figure is represented in a small region. SAM Adapter and AutoSAM provide a multi-organ segmentation with noise. While SAMed outputs a result with lesser noise, it is also confused by the shape of the organ. H-SAM outputs a perfect result with the correct shape and no noise.

\subsection{Visualization on Synapse dataset}
\begin{figure*}
  \centering
   \includegraphics[width=0.8\linewidth]{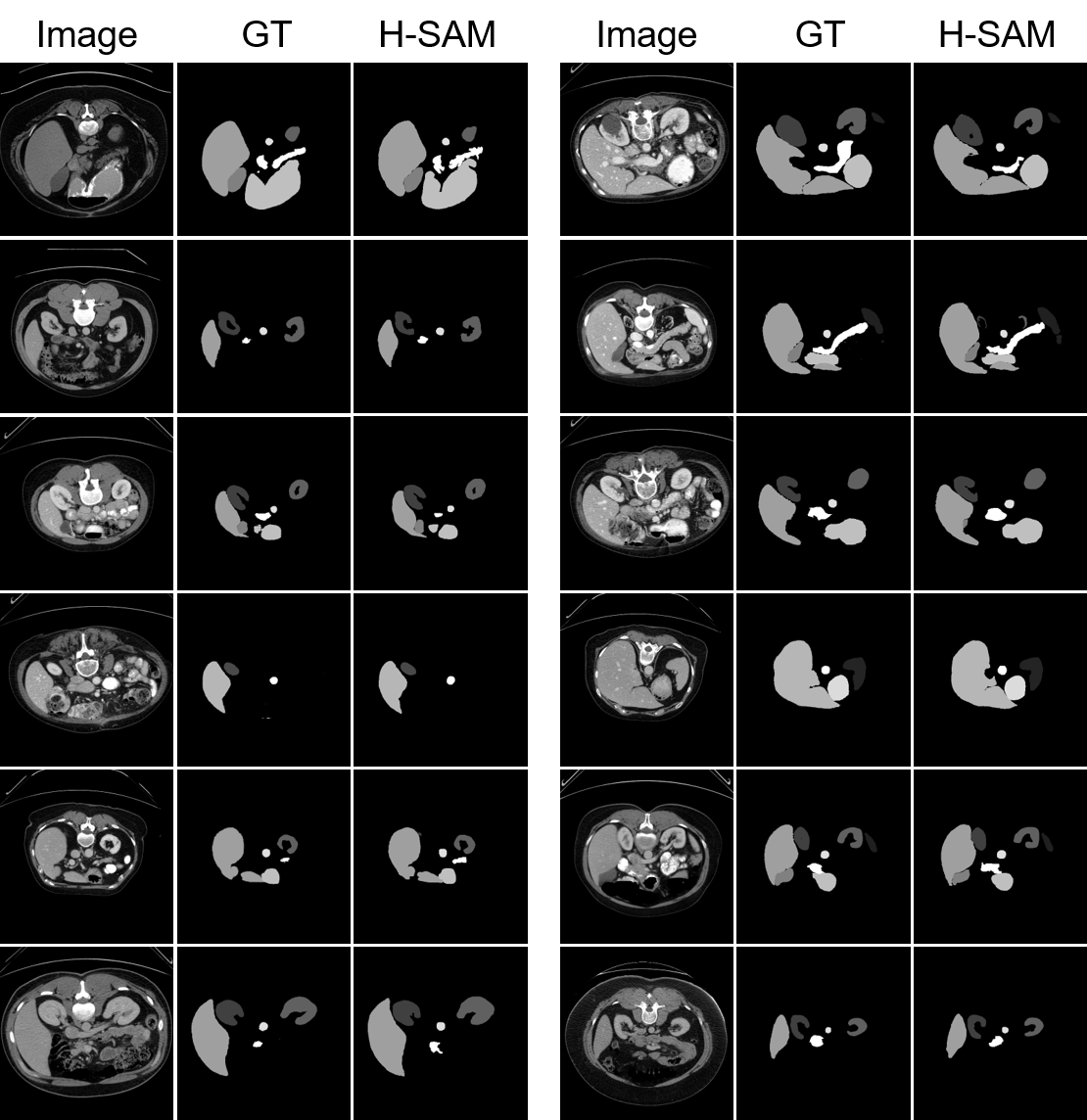}

   \caption{Visualization of semantic segmentation predictions on the Synapse dataset. First and fourth columns: raw image. Second and fifth columns: ground truth. Third and sixth columns: prediction.}
   \label{supp1}
\end{figure*}

As shown in Figure~\ref{supp1}, we present the visualization of semantic segmentation predictions on the Synapse dataset. Compared to ground truth, H-SAM performs promising results with both multiple organs (up to 8) and fewer organs.

\subsection{Visualization on 2 stages}
\begin{figure*}
  \centering
   \includegraphics[width=0.8\linewidth]{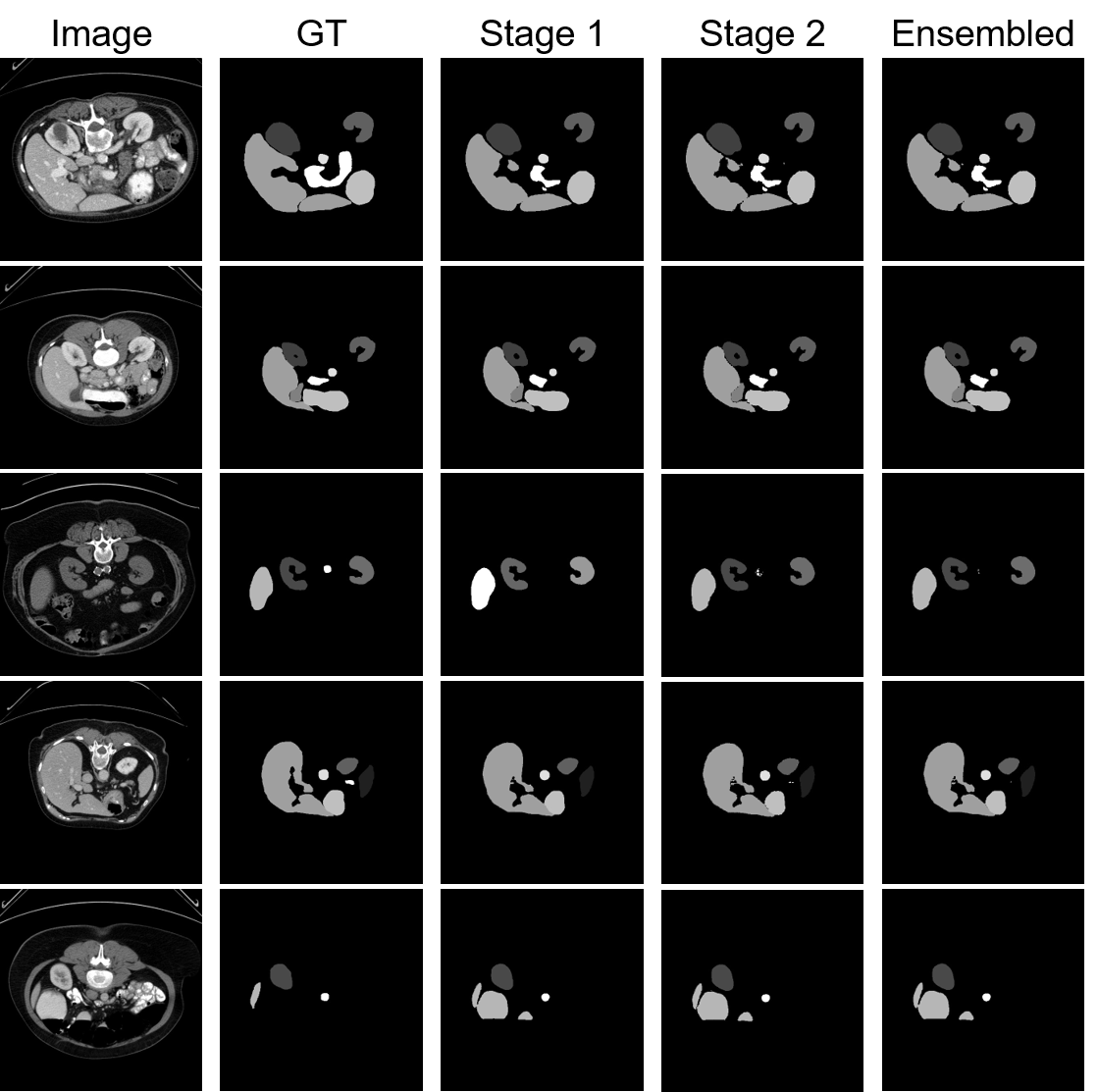}

   \caption{ Visualization of the outputs from different stages. First column: raw image. Second column: ground truth. Third column: stage-1 output. Fourth column: stage-2 output. Fifth column: ensembled output from 2-stage outputs. The last row shows failure cases where stage 2 takes an erroneous prediction from stage 1 as the prior.}
   \label{stage}
\end{figure*}

Here we present the visualization of outputs from different stages. As shown in Figure~\ref{stage}, benefit from the joint training design, both of the 2 stages perform excellent segmentation predictions. In row 5, we present a failure case where stage 2 takes an erroneous prediction from stage 1 as the prior and mistakes a background region to kidney. However, in most cases, the stage-2 prediction takes and corrects stage-1 results as the prior to generating finer segmentation, which especially can be reflected from small organs like Pancreas, as shown in rows 3 and 4. 
\vspace{8cm}

  \end{appendices}

% WARNING: do not forget to delete the supplementary pages from your submission 
% \input{sec/X_suppl}

\end{document}